\documentclass[final]{cvpr}

\usepackage{times}
\usepackage{epsfig}
\usepackage{graphicx}
\usepackage{amsmath}
\usepackage{amssymb}
\usepackage{algorithm}
\usepackage{algorithmicx}
\usepackage[noend]{algpseudocode}

\usepackage{tabularx}
\usepackage{adjustbox}
\usepackage{threeparttable}
\usepackage{multirow}
\usepackage{subfigure}
\usepackage{caption}

\usepackage{graphicx}
\usepackage{amsmath,amssymb} %
\usepackage{color}

\usepackage{booktabs}

\usepackage[breaklinks=true,bookmarks=false]{hyperref}

\DeclareMathOperator*{\argmax}{argmax}

\newcommand{\Supp}{Appendix}

\begin{document}

\title{Diverse Complexity Measures for Dataset Curation in Self-driving}

\author{Abbas Sadat$^1$, Sean Segal$^{1 2}$, Sergio Casas$^{1 2}$, James Tu$^1$ \\Bin Yang$^{1 2}$, Raquel Urtasun$^{1 2}$, Ersin Yumer$^1$\\
Uber ATG$^{1}$, University of Toronto$^{2}$\\
{\tt\small \{abbas.sadat, meyumer\}@gmail.com}\\
{\tt\small \{seansegal, sergio, byang, urtasun\}@cs.toronto.edu james.tu@mail.utoronto.ca} 
}

\definecolor{raquel_colour}{RGB}{255,0,0}     
\newcommand{\raquel}[1]{{\color{raquel_colour} Raquel: #1}}

\maketitle

\begin{abstract}
   
Modern self-driving  autonomy systems heavily rely on deep learning. As a consequence, their performance  is influenced significantly by the quality and richness of the training data. Data collecting platforms can generate many hours of raw data in a daily basis, however, it is not feasible to label everything. It is thus of key importance to have a mechanism to identify "what to label". Active learning approaches  identify  examples to label, but their interestingness is tied to a fixed model performing a particular task. These assumptions are not valid  in self-driving, where  we have to solve a diverse set of  tasks (i.e., perception, and motion forecasting) and our models  evolve over time frequently. In this paper we introduce a novel approach and propose a new data selection method that exploits a diverse set of criteria that quantize interestingness of traffic scenes. Our experiments on a wide range of tasks and models show that the proposed curation pipeline is able to select datasets that lead to better generalization and higher performance. 
\end{abstract}


\section{Introduction}

Self-driving has recently benefited from deep learning breakthroughs, which have enhanced  the  performance of autonomy systems significantly. 
The performance achieved by these systems is tightly coupled to the quality, size and richness of training datasets. Furthermore, as self-driving is a safety critical application it is very important to have a diverse set of testing scenarios that are representative of driving. 

Collecting data is a fairly easy process -- a single vehicle can generate several Tb of data a day. 
However, it is not feasible to label everything that has been collected. For instance, it could cost around \$150K\footnote{scale.com} to simply annotate the bounding-box of objects in one hour of camera data, assuming average density of 50 objects per image.
Hence, it is of key importance to have a mechanism to identify "what to label" such that we can get the most relevant labeled dataset to achieve the highest autonomy performance given a labeling budget.

One of the most relevant areas of research in this spirit is active learning, which has been used as a mechanism to identify interesting examples to label. However, active learning approaches are  tied to a  model solving a given task\footnote{We consider an ensemble to be a model belonging to a specific model class}, while in our setting many tasks need to be performed -- a self-driving car needs to perceive the world, predict the future trajectory of all the actors in the scene and perform safe motion planning. 
Furthermore, the assumption in active learning approaches is that the model is fixed and we are interested in improving its performance via additional labels. 
However, in current modern self-driving approaches the autonomy stack is constantly changing, and  as a consequence, examples that might be informative to label a few weeks ago might not be interesting anymore under the evolution of the perception, and motion forecasting modules. 
As a consequence, existing self-driving benchmarks have not been created by using active learning, but instead by exploiting random sampling, handcrafted set of heuristics, or manual selection. 

In this paper we look at this task with a new lens and define specific criteria to quantize interestingness in order to identify a dataset of challenging and diverse scenarios for self-driving tasks. These criteria are not bound to a specific autonomy architecture or model and does not require multiple iterations of training and evaluation. 
More specifically, we propose a set of complexity measures to characterize self-driving scenarios. 
These measures include various factors such as the map, static and dynamic objects surrounding the ego-vehicle, and the executed trajectory of  the data collection platform  as well as the possible interactions it performs with other road-users. 
We then utilize these complexity measures inside a  novel  data selection approach  to curate challenging and diverse dataset of traffic scenes.
Through extensive experiments using various models in the literature, that such curation approach leads  to better generalization of a wide variety of models in different autonomy tasks. More specifically, we show average 6.5\% improvement in performance of various perception models and average 6\% improvement in motion forecasting models from the recent state-of-the-art compared to other approaches such as active learning. 

\section{Related Work}

\subsection{Data Selection}
Finding hard examples has been used to train models more effectively. 
In \cite{felzenszwalb2009object} an active training set is maintained and is iteratively updated during training by removing easy examples and adding harder ones based on, e.g.~, classification margins. 
In \cite{shrivastava2016training} hard examples are mined online at ROI level for object detection task. However such methods deal with data that is already labeled.

Data selection has been studied in the \textit{active learning} literature where batches of unlabeled data are selected for labeling by iteratively training a model on the labeled set and finding informative unlabeled data. 
Uncertainty-based methods select difficult examples by considering entropy in predicted distributions \cite{joshi2009multi, li2013adaptive}, uncertainty across a model ensemble \cite{beluch2018power, gal2017deep} or the estimated loss of each example \cite{yoo2019learning}.
To avoid training a full model multiple times, \cite{coleman2019selection} proposed to use a simpler model as a proxy to perform efficient data selection. 
The proxy model can be either a simpler architecture or the original model trained with fewer iterations. Diversity-based approaches \cite{nguyen2004active, Sener2018ActiveLF} aim to select a subset of the unlabeled pool that best represents the entire dataset. \cite{haussmann2020scalable} describes an scalable production system for active learning in object detection where various scoring and sampling strategies are compared. 

\textit{Semi-supervised} active learning frameworks has also been proposed recently. In \cite{simeoni2019rethinking} labeled data as well as the unlabeled ones are used during model training in each active learning cycle, for example by treating the most confident prediction for unlabeled data as pseudo label. 
Similarly in \cite{gao2019consistency} the model is trained with a usual loss on labeled data (e.g. cross-entropy) and a consistency loss on unlabeled data which penalize highly inconsistent predictions of slightly distorted samples  of an example data.

Active learning and semi-supervised approaches above assume that the model is fixed and we are interested in improving its performance via additional labels. 
However, our work does not make this assumption since examples that might be informative to label a few weeks ago might not be interesting anymore under the evolution of the different self-driving stack modules. 

\subsection{Self-driving Datasets}
Many self-driving datasets have been released publicly in the past few years. 
Kitti \cite{geiger2012we}, most notably, was the first benchmark with a dataset that included multiple sensors of LiDAR and camera supporting the tasks of stereo, optical flow, visual odometry, and 3D object detection. 
Citiscapes dataset \cite{cordts2016cityscapes} includes annotated images from various cities in multiple seasons, suitable for semantic and instance segmentation. 
BDD100K dataset \cite{yu2020bdd100k} offers a diverse set of images that has been crowd-sourced with annotations such as 2D labels, lane markers and weather. 

LiDAR sensor data particularly has rich 3D information making such datasets suitable for tasks of 3D detection and possibly tracking and motion forecasting \cite{wang2019apolloscape, geyer2020a2d2}. 
Waymo open dataset \cite{sun2019scalability} includes 2D/3D bounding boxes with camera and LiDAR data collected over large geographical and time of day extent. Argoverse \cite{chang2019argoverse} also provides map rasters and graph with lane and intersection annotations. 
In order to create an interesting and challenging benchmark for trajectory forecasting task, they mine vehicles that are at intersection, performing a turn action or in dense traffic. 
nuSenes \cite{caesar2020nuscenes} is the first dataset that includes radar data and supports tasks of 3D detection and tracking as well as behavior prediction. 
The scene selection is achieved manually to include dense traffic, rare classes, dangerous traffic situations, and difficult AV maneuvers.
Other datasets include only sensor data without any human annotations of objects \cite{santana2016learning, agarwal2020ford}.

In contrast to the above mentioned works, we propose a method that combines a systematic semantic analysis of the raw data from all aspects relevant to driving:  the complexity of the map , other actor motion and configurations, and last but not least the SDV pose and motion. We utilize these complexity measures in a framework that encourages not only challenging but a diverse selection to arrive at our curated dataset. 


\section{Data Selection via Complexity Measures}

In this section we propose a novel approach to decide which frames are more interesting to be labeled in order to create a rich and diverse dataset for training and evaluation. 
Towards this goal, we define a number of complexity measures driven by the static constructs in a driving scenario (e.g.,  
lane topology, presence of an intersection), the traffic participants (e.g., other vehicles, pedestrians) and the maneuvers that the ego-car performs. 
Note that our data collection platform is a self-driving vehicle (SDV) and thus we will use these two terms interchangeably. 

We assume the existence of a prebuilt HD Map for the areas that we will drive, a  localization system that can localize the vehicle with respect to this map, as well as a 
perception software stack that can be ran on the collected raw data  whose by product are 
 object detections for each frame  and their associated tracks that link those detections across time. 
 
 In the following, we first introduce our method to select challenging and diverse scenarios given a set of complexity measures. 
 We then discuss in detail the complexity measures we exploit in our work.

\subsection{Data Selection}
Our selection process operates at the snippet level (\ie a sequence of frames) instead of frame level. 
This allows observing phenomena that have a temporal component such as a lane-change or an interaction between actors, or the change in the speed of the SDV. 
Moreover, labeling a sequence of consecutive sensor data can be done more efficiently compared to to scattered individual frames.
More formally, we define a snippet
$s=\left\{f_i|i=0,...,T-1\right\}$ as a set of  $T$ consecutive frames of data. In our experiments, each snippet represents 25 seconds of data with 250 frames.
Throughout this paper, we will interchangeably refer to a snippet as a scenario.
The purpose of the data selection process is then to pick $K$ non-overlapping snippets from the pool of unlabeled data. 

We formulate the snippet selection as an optimization problem where first interesting scenarios are selected, and then the dataset is enhanced to be diverse and complete. 

\paragraph{Selecting Challenging Scenarios:}
In order to identify a set of challenging scenarios, we first rank the snippets using a scoring function $g$. 
In particular, we utilize a linear combination of the complexity measures as our "interestingness" score for a given snippet $s$
$$g(s;w)=w^TE(s)$$  
where $E(s)$ is the vector of complexity measures.

It is important to mention that   what is considered to be challenging/interesting depends on the target task. F
or example, imagine a scenario where the SDV is stopped at a busy intersection with a red light. 
As there are many interactions  between actors, this is an interesting  scenario for the tasks of perception and motion forecasting, however,  for  motion  planning, this scenario may not be very useful as the SDV is not moving. 
On the other hand, there can be many scenarios where the SDV needs to interact with one or two actors of interest in an empty intersection, making this scenario  interesting  for  planning and motion forecasting tasks, but not for detection. 
Therefore, we use different weight vector $w$ when ranking the snippets for each specific task.

We assume there are $n$ tasks and the goal is to select $c_i, i=1,...,n$ snippets for each task $i$. Therefore we will have the following optimization:
\begin{align}
    \label{eq:opt}
    S^*_1, ..., S^*_n = \argmax_{\substack{S_1, ..., S_n \subset \Psi \\ |S_i| = c_i, \forall i \\ s_i \cap s_j = \emptyset, \forall s_i,s_j \in \cup_i S_i}}  \sum_i^n \sum_{s \in S_i} g(s;w_i)
\end{align}
where $\Psi$ is the set of all snippets, $w_i$ corresponds to the complexity weight vector specific for task $i$. 
The constraints simply makes sure there are no overlapping snippet within or across selected sets. 
Note that for our experiments, we consider two tasks: perception and motion forecasting (i.e. prediction). 
However, we still consider motion planning when we consider the distribution of the data, since as the downstream task it will have indirect affect on both perception and motion forecasting performance. 
We choose the weight vector for each task is tuned empirically.

We solve this optimization problem by first ranking all the snippets and then  iteratively picking the next most "interesting" snippet from the queue  and removing any overlapping one from the candidate set. 
We repeat this process for a fix number of iterations until we have reached a fixed budget.

\paragraph{Selecting Diverse Scenarios:}
Limiting the data selection to only challenging scenarios will not necessarily lead to a diverse dataset, or to a complete set of scenarios that we might encounter in the real world. The goal of this additional  selection step is to identify a set of snippets that  ensures completeness and diversity. 
We quantize the dissimilarity between snippets as a function of their difference in the complexity measures, where in  order to get geo-diversity we expand the complexity vectors with the latitude and longitude coordinates of the frames. 
We then iteratively look for the snippet that is farthest from the current selected set and added to the set to be labeled. In particular, at each iteration we select

\begin{equation}
    \label{eq:diversity}
    s^* = \argmax_{s\in\{\Psi-\mathcal{S}\}}  \bigl( \min_{s^\prime \in \mathcal{S}} d(s, s^\prime)\bigl),
\end{equation}
where $\Psi$ is the set of all unlabeled snippets, $\mathcal{S}$ is the set of already selected snippets, and $d$ is a dissimilarity  function. 

In order to capture diversity at a granular level, we have define the maximum difference between the closest pair, as our  dissimilarity function between snippets

\begin{equation}
d(s_i, s_j) = \max_{k} \min_{l} ||E(s_i,k)-E(s_j,l)^\prime||_2
\end{equation}
where $k$, $l$ index over the frames of the $s_i$ and $s_j$ snippets respectively. 
The the full process of data selection is depicted in Algorithm \ref{alg:selection}.

\begin{algorithm}[t!]\small
    \caption{Data Selection}
    \label{alg:selection}
    \begin{algorithmic}[1] 
        \Procedure{Select}{$\Psi$, $E$, weight vectors $w_i$ for each task, desired number of snippets $k_i$ for each task, desired number of diverse snippets $k_{div}$}
            \State $S_1, ..., S_k$ $\gets \emptyset$
            \Statex \Comment{Select challenging scnarios}
            \While{$|S_i| < k_i, \exists i$}
                \For{ each task $j$}
                    \If{$|S_j| < k_j$}
                        \State $s^* \gets \argmax_{s \in \Psi-\cup_i S_i} g(s; w_j)$
                        \State $S_j \gets S_j \cup \{s^*\}$
                    \EndIf
                \EndFor
            \EndWhile
            
            \Statex \Comment{Select diverse scnarios}
            \State $S \gets S_1 \cup, ..., \cup S_k$
            \State $S_{div} \gets \emptyset$
            \For{$i$ =$1, ..., k_{div}$}
                \State $s^* \gets \argmax_{s \in \Psi-S\cup S_{div}} \min_{s^\prime \in S\cup S_{div}} d(s, s^{\prime})$
                \State $S_{div} \gets S_{div} \cup \{s^*\}$
            \EndFor
            \Return $S_1, ..., S_n, S_{div}$

        \EndProcedure
    \end{algorithmic}
\end{algorithm}

\subsection{Complexity Measures}
In order to characterize a traffic scene, we consider the map surrounding the location of the SDV and the detected objects within a region of intersect (ROI) around  it. 
We also consider elements from the scenario that are directly related to the SDV maneuver. 
\begin{figure*}
	\centering
	\includegraphics[width=0.90\textwidth]{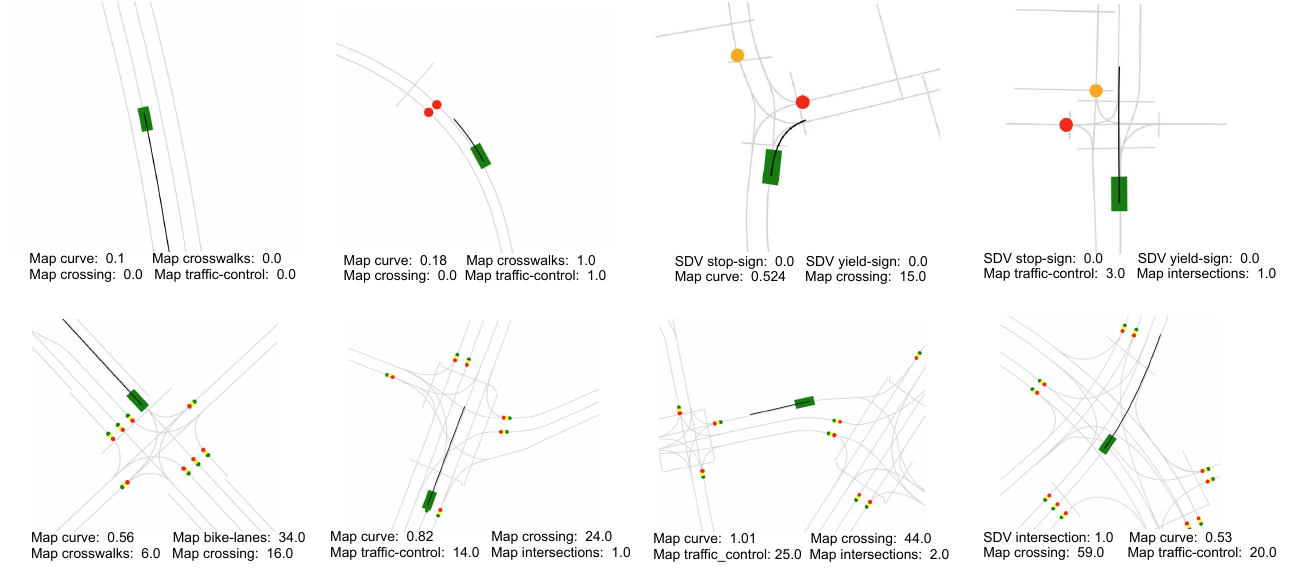}
	\caption[]
    {Examples of the infrastructure-related complexity measures. In the top row, the path complexity measure of the lanes increases, from zero curvature to large varying curvature. in the bottom row, the map crossing measures indicates the complexity of the lane-graph topology as can be seen in the scene visualizations.}
    \label{fig:map_comp}
\end{figure*}
\subsubsection{Infrastructure-related Complexity Measures}
We utilize a diverse set of complexity measures related to the static part of the environment. 
This contains information about how the vehicles might drive in the scene, the presence of intersections as well as traffic control elements and other road elements such as bike lanes and crosswalk. 

\paragraph{Geometry and topology of driving paths:}
We define \textit{driving paths} as a plane curve in $\mathbb{R}^2$, representing the center-line of the map lanes. 
A driving path of constant curvature is a straight line or a circle and a vehicle can follow such path simply with a constant steering wheel. On the other hand, paths with variable curvature require more complex steering as shown in Figure \ref{fig:map_comp}. 
We use this intuition to define the complexity of a path. 
Specifically, we represent a path, $C(s)$, as a finite set of $K$ way points sampled along its arc-length, 
$s$: $\mathcal{C} = \left\{C(s_i) | 0 \leq s \leq 1, i=0,...,K\right\}$. 
Using a finite-difference method, the curvature (and rate of change in curvature) can be computed for each way point, resulting in the 
set: $\mathcal{K}_\mathcal{C}=\left\{\kappa(s_i) | 0 \leq s \leq 1, i=0,...,K\right\}$. 
Finally, we propose to use the mean of curvature $\mu(\mathcal{K}_\mathcal{C})$, and the mean of its derivative, $\mu(\dot{\mathcal{K}}_\mathcal{C})$, 
as the measure of complexity of the curve: 
\begin{equation}
\label{eq:curve_complexity}
E^{\text{curve}} = \mu(\mathcal{K}_\mathcal{C}) + \mu(\dot{\mathcal{K}}_\mathcal{C}),
\end{equation}
The driving paths in the map can cross each other creating scenes where vehicles can have potentially conflicting goals, and hence interesting. We measure such complexity by $E^{\text{crossing}}=\sum_c v_c$ as with $v_c$ being the number of times a driving path $c$ is crossed by other lanes. Figure \ref{fig:map_comp} shows various examples of lanes and their complexity measures.

\paragraph{Intersections, traffic-lights, and signage:}
Traffic scenes are generally more interesting at intersections. 
Hence we consider whether the SDV is at an intersections or not, as well as the complexity of the topology of the intersection by counting  the number of roads reaching the intersections and the number of lanes that exist at each road. 
Besides, we also compute the number of traffic-lights and other signage such as stop-signs or yield-signs. 

\paragraph{Bike-lanes and crosswalks:}
The existence of bike-lanes and how they interact with vehicle lanes can increase the complexity of a traffic scene. 
To capture this we use the same vehicle-lane geometry and topology measures mentioned above for bike-lanes. 
Similarly, we consider crosswalks and how they expand over the vehicle lanes.

\paragraph{Drivable-area height variation:}
Driving on hilly areas can be more challenging compared to flat roads. 
We define an additional complexity as the height variance in the drivable surface, \ie $E^{\text{height}} = \sigma^2(\mathcal{Z})$ where $\mathcal{Z}$ is a set representing the map height of points uniformly sampled on lanes.

\subsubsection{Traffic Participants}
Other important aspects of a scenario are how crowded the scene is as well as the diversity of its actors. 
We thus define the following complexity measures. 

\paragraph{Crowdedness:}
We measure the number of objects  in an ROI around the SDV to capture how crowded a traffic scene is, using the following:
\begin{equation}
    \label{eq:crowd}
    E^{\text{crowd}} = \frac{1}{T}\sum_{t}^T|\mathcal{D}_{t}|,
    \end{equation}
where $\mathcal{D}_{t}$ is the set of detections in frame $f_t$.
We measure this separately for static and dynamic actors to have more granular information. 
Note that this does not measure how the object can potentially interact with SDV which will be covered by different complexity measures.

\paragraph{Class and spatial diversity:}
Many interesting interactions can happen when there are multiple types of actors in a traffic scene. 
Some classes of actors (e.g. bicyclists) are orders of magnitude more rare than others (e.g., vehicles). 
We thus measure the diversity of actors by:
\begin{equation}
    \label{eq:class}
    E^{\text{class}} = \frac{1}{T}\sum_{t}\frac{1}{|\mathcal{D}_{t}|}\prod_c\left(1+|{}_c\mathcal{D}_{t}|\right),
    \end{equation}
where ${}_c\mathcal{D}_{t}$ is the set of detections that belong to class $c$ in frame $f_t$. Figure \ref{fig:actor_sdv_comp} shows various traffic scenes and their actor class diversity measure.
In addition we measure the variance of the distance to the SDV for those actors.

\paragraph{Path and speed diversity:}
Up to now we have introduced measures related to the existence of certain actors. 
However, it is important to take into account how those actors move. 
Similar to the complexity of the driving-paths in (\ref{eq:curve_complexity}), we use the curvature of the path that each actor took, along with its first derivative to measure the complexity of the actor's behavior. 
This measure can capture many interesting events. 
For example a vehicle that is making a lane-change follows a path with high curvature change. Similarly, pedestrians the change the direction of their motion will lead to high path complexity. 
Such behaviors of actors will serve as rich examples for training prediction models.

The variation in the speed of actors can also add to the complexity of the traffic scene indicating an interesting interaction of an actor with another one, a traffic-control element, or can simply show an intention to change path. We define a measures reflecting the speed variance of an actor as well as the variance of the mean speed of all actors in a given scene:
\begin{equation}
    \label{eq:speed}
    E^{\text{speed}} = \sigma^2(\Omega)+\sum_i^{|\Omega|}\sigma_i^2(\omega_i),
\end{equation}
where $\omega_i$ is a discreet speeds computed for the $i^{\text{th}}$ actor and $\Omega$ is the set of average speeds for all the actors.

\subsubsection{SDV Maneuvers}

Finally it is important to identify scenarios where the SDV performs complex maneuvers. 
We thus introduce the following measures to capture how interesting a scenario is as it relates to the SDV.

\begin{figure*}
	\centering
	\includegraphics[width=0.95\textwidth]{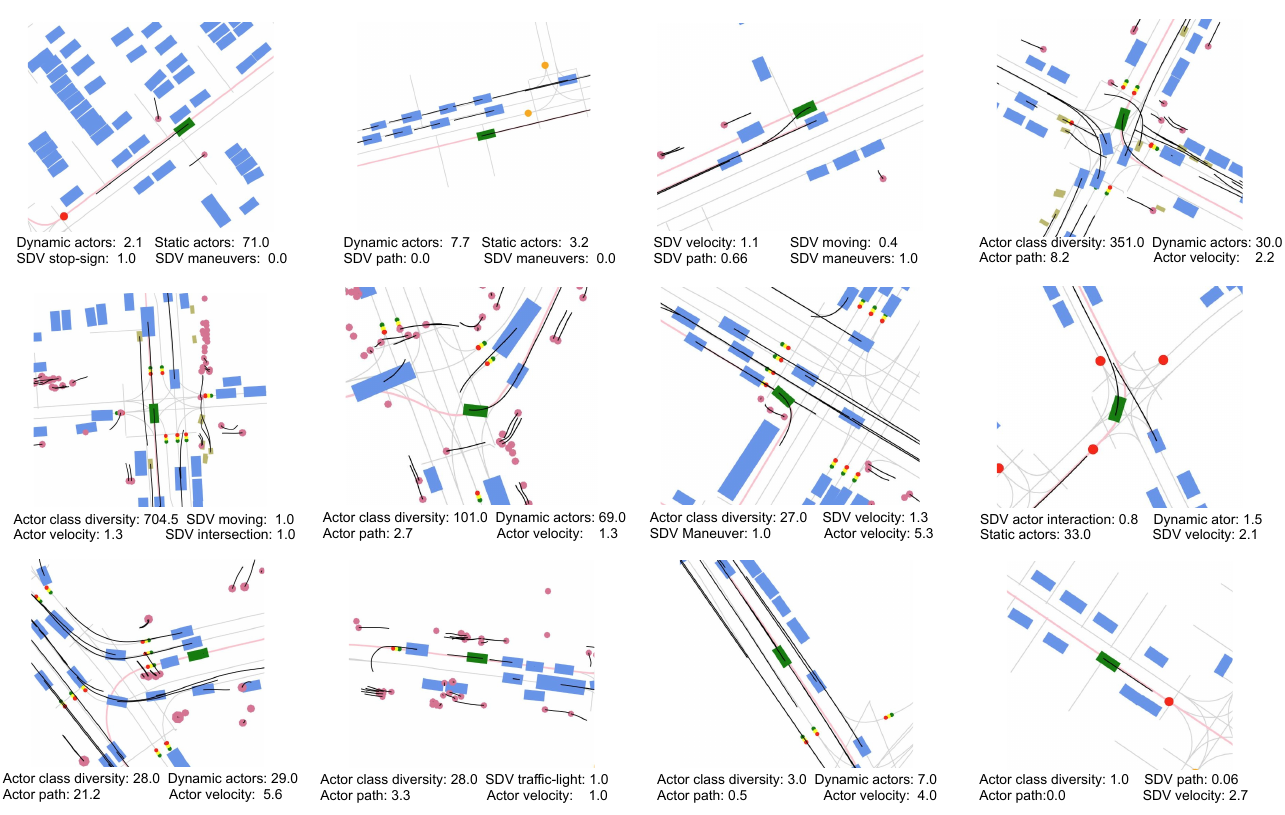}
	\caption[]
    {Examples of complexity measures related to the traffic participants and the SDV maneuver. The top row shows compares, for example, the static-crowd measure, increasing from left to right, v.s.~ the dynamic-crowd complexity measure.  In the middle row, the class diversity measure increases from right to left, increasing the complexity of the traffic scene. It also indicates how the complexity of SDV path corresponds to the complexity of the scene. Lastly, in the bottom row the path  and velocity taken by the actors become more complex from right to left, leading to increasingly more challenging scenarios for tasks such as motion forecasting.}
    \label{fig:actor_sdv_comp}
\end{figure*}

\paragraph{Path and speed:}
We use the same measures of (\ref{eq:curve_complexity}, \ref{eq:speed}) introduced for actors to obtain the complexity of the ego vehicle path and motion.
Figure \ref{fig:actor_sdv_comp} shows various SDV trajectories and their speed and path complexities.

\paragraph{Route:}
Some of the the high-level maneuvers of the ego vehicle are naturally less frequent that others, e.g. lane-change and turns \textit{v.s} lane-following.
Therefore, we count such maneuvers to measure the complexity of the SDV route.
We also consider whether SDV interacts with traffic-lights or stop-signs while following the route.

\paragraph{Interactions with other actors:} We obtain the number of static and dynamic objects that are within distance to the SDV path as measure of how other objects affect SDV behavior.
However distance is not always a sufficient measure. 
For example, vehicles that are waiting to make an unprotected left turns, or passing through and intersection controlled by stop-sign, may not be close to the ego path, but their behaviors affect SDV's decision and vice versa. 
To capture this, we measure the number of vehicles that pass through a lane that is in conflict with SDV route. 
We also count the vehicles that can reach such lanes within a short time horizon to capture potential actors interacting with SDV.
Figure \ref{fig:actor_sdv_comp} shows an example of SDV interacting with another vehicle at an intersection.

\paragraph{Nudging maneuvers:}
In some scenarios the SDV needs to partially move out of its lane and come back in order to pass an object or vehicle that is partially blocking the path. We specifically capture these maneuvers as they make the scenario very complex.

\section{Experiments}
\paragraph{Experimental Setup:}
We use a very large labeled dataset that consists of 140 hours of manual driving data (20k+ snippets of 25s). 
The dataset is split into test and validation sets as well as a base training set $\Psi$, ensuring no geographical overlap and similar label distributions. 
Using our proposed approach (CR), random sampling (RN), and an active learning baseline (AL), we select two separate training sets of size 1,000 and 3,000 snippets from $\Psi$. 
Similarly, we we select 2 test sets out of the original test set: \textit{Easy} which is selected randomly, and \textit{Hard}, which is selected using our proposed approach.
We train various perception and motion-forecasting models using the curated training sets, and evaluate them on the two test sets.
Specifically for perception-only task we use PointPillars\cite{lang2019pointpillars}, and PointRCNN\cite{shi2019pointrcnn}. 
Additionally, we use joint perception-prediction models of ILVM\cite{casas2020implicit}, MultiPath\cite{chai2019multipath}, and ESP\cite{rhinehart2019precog} in which both tasks are trained end-to-end. 

\paragraph{Metrics:} For the perception task, we use mean average precision (mAP) to compare models. Note that this metrics is computed for each class of \textit{Vehicles}, \textit{Bicycles}, and \textit{Pedestrians}. 
For motion forecasting we use the following metrics: mean average displacement error (meanADE), minimum average displacement error (minADE), and collision rate among actors. 
The motion prediction metrics are computed over 5s horizon, similar to how they are trained.
Following \cite{casas2020implicit}, we evaluate prediction in true positive object detections at a common recall point across models. That means, we find the detection threshold for each model such that all of them are evaluated at the same recall point of the detection Precision-Recall curve.

\paragraph{Active Learning Baseline (AL):}
We compare our dataset curation method against an uncertainty-based active learning approach. More specifically, we select snippets with high entropy predictions generated by a trained prediction model.  
We use a prediction model with the same backbone as ILVM \cite{casas2020implicit}. 
However, in order to compute entropy easily, the header is replaced with a simplified model that outputs a distribution $p(y)$ with an independent 2D Gaussian for each actor $i$ and time-step $t$, 
\begin{equation}
	p(y) = \prod_{i}^{N} \prod_{t}^{T} p(y^{i}_{t}; \mu^{i}_{t}, \Sigma^{i}_{t})~~.
\end{equation}

\noindent Afterwards, for each frame, we can compute the entropy of the output distribution as, 
\begin{equation}
	H(y) = \sum_{i}^{N} \sum_{t}^{T} \frac{1}{2} \log \lvert 2 \pi e \Sigma^{i}_{t} \rvert~~,
\end{equation}
and sum the entropies across all frames in a snippet to obtain a final uncertainty score. 
Given the size of the base set $\Psi$, it is infeasible to iteratively re-train and re-score all examples more than once.  
Therefore, we train the prediction model initially on a random subset of $250$ snippets and then select the remaining snippets with the highest entropies to obtain datasets of size 1k and 3k snippets. 
Specific model details are available in the supplementary materials.

\subsection{Results}
Table \ref{tbl:sum} shows the summary of all results where the metrics are averaged per model and object classes (macro-averaging). 
It is clear that our curation method outperforms the baselines on average in all tasks and training-set sizes. 
Specifically, in object detection we improve 7.5\% and 6.3\% in the \textit{Easy} set over active learning respectively in 1k and 3k settings, and 7.8\% and 6.7\% in the \textit{Hard} set. 
A similar trend is observed in the motion forecasting task. 
Compared to active learning, our approach gains 10.3\% and 8\% improvement in prediction metrics for 1k and 3k settings in the \textit{Easy} test set, and 6.3\% and 4\% in the \textit{Hard} set. 
The results indicate that using the data selected by our proposed method, we can achieve significantly better generalizations in various models and tasks in self-driving. 
In the following sections we present the result of each model for each task.

\begin{table}[t]
	\centering
		\begin{threeparttable}
			\begin{tabularx}{0.49\textwidth}{
				>{\centering\arraybackslash}X |
				>{\centering\arraybackslash}X >{\centering\arraybackslash}X >{\centering\arraybackslash}X>{\centering\arraybackslash}X >{\centering\arraybackslash}X}
			\toprule
			Dataset size   & Selection Method &  \multicolumn{2}{c}{Detection $\uparrow$}  & \multicolumn{2}{c}{Prediction $\downarrow$}\\
						   &				  &  Easy & Hard 					&		Easy & Hard \\
			\midrule
			\multirow{3}{*}{1k} &RN&0.543&	0.545&0.85&	0.95\\
								&AL&0.532&	0.545&0.77&	0.79\\
								&CR&\textbf{0.575}&	\textbf{0.588}&\textbf{0.69}&\textbf{0.74}\\
            \midrule
            \multirow{3}{*}{3k} &RN&0.594&	0.596&0.81&	0.83\\
								&AL&0.587&	0.589&0.75&	0.74\\
								&CR&\textbf{0.624}&	\textbf{0.629}&\textbf{0.69}&	\textbf{0.71}\\

			\bottomrule
			\end{tabularx}
		\end{threeparttable}
    \caption{Detection and Prediction results averaged over the models and actor classes.}
    \label{tbl:sum}
\end{table}
\paragraph{Detection:}
Table \ref{tbl:detection} shows the detection metrics for detection-only models a well as joint perception-prediction models.
The results indicate that our approach consistently improves Bicycle detections significantly across all models and training set sizes. 
Similarly for Pedestrians, our approach leads to better mAP in majority of the models. 
Interestingly for Vehicle detection both our approach and random sampling show strong performance in the \textit{Hard} test set across different models.
\begin{table*}[t]
	\centering
		\scriptsize
		\begin{threeparttable}
			\begin{tabularx}{\textwidth}{
				>{\centering\arraybackslash}c |  %
				>{\centering\arraybackslash}X |
				>{\centering\arraybackslash}X |
				>{\centering\arraybackslash}X >{\centering\arraybackslash}X >{\centering\arraybackslash}X
				>{\centering\arraybackslash}X >{\centering\arraybackslash}X >{\centering\arraybackslash}X}
			\toprule
			Model 	              &  Size & Train set  & \multicolumn{2}{c}{Vehicle mAP@0.7}  & \multicolumn{2}{c}{Bicycle mAP@0.3} & \multicolumn{2}{c}{Pedestrian mAP@0.3}\\
			 &&&Easy&Hard&Easy&Hard&Easy&Hard\\

			 \multirow{6}{*}{PointPillars\cite{lang2019pointpillars}} 		  &\multirow{3}{*}{1k}&RN&\textbf{0.817}&\textbf{0.812}&0.224&0.222&0.665&0.649\\
			 &&AL&0.806&0.805&0.158&0.175&\textbf{0.695}&0.727\\
			 &&CR&0.802&0.802&\textbf{0.234}&\textbf{0.246}&0.691&\textbf{0.729}\\
			 \cline{3-9}

&\multirow{3}{*}{3k}&RN&\textbf{0.851}&\textbf{0.847}&0.328&0.324&0.753&0.772\\
		 &&AL&0.840&0.838&0.276&0.268&0.752&0.776\\
		 &&CR&0.846&0.845&\textbf{0.358}&\textbf{0.365}&\textbf{0.763}&\textbf{0.785}\\
\midrule
\multirow{6}{*}{PointRCNN\cite{shi2019pointrcnn}} 		  &\multirow{3}{*}{1k}&RN&\textbf{0.733}&\textbf{0.731}&0.321&0.323&0.645&0.674\\
			 &&AL&0.707&0.706&0.209&0.289&0.662&0.690\\
			 &&CR&0.731&\textbf{0.731}&\textbf{0.392}&\textbf{0.395}&\textbf{0.680}&\textbf{0.708}\\
			 \cline{3-9}

&\multirow{3}{*}{3k}&RN&\textbf{0.746}&0.744&0.402&0.400&0.677&0.701\\
			 &&AL&0.724&0.722&0.369&0.362&0.688&0.714\\
			 &&CR&\textbf{0.746}&\textbf{0.746}&\textbf{0.448}&\textbf{0.452}&\textbf{0.695}&\textbf{0.721}\\		
			 
			\midrule
			\multirow{6}{*}{ILVM\cite{casas2020implicit}} 		  &\multirow{3}{*}{1k}&RN&\textbf{0.756}&\textbf{0.735}&0.183&0.172&0.671&0.676\\
																	&&AL&0.732&0.709&0.159&0.168&0.685&0.684\\
																	&&CR&0.744&0.733&\textbf{0.251}&\textbf{0.259}&\textbf{0.690}&\textbf{0.693}\\
																	\cline{3-9}
																	&\multirow{3}{*}{3k}&RN&0.782&0.757&0.267&0.257&0.706&0.708\\
																					   &&AL&0.789&0.777&0.272&0.275&\textbf{0.737}&\textbf{0.731}\\
																	&&CR&\textbf{0.795}&\textbf{0.781}&\textbf{0.313}&\textbf{0.334}&0.727& 0.725\\

			\midrule
			\multirow{6}{*}{MultiPath\cite{chai2019multipath}} 		  &\multirow{3}{*}{1k}&RN&\textbf{0.720}&0.719&0.160&0.198&0.676&0.689\\
																						 &&AL&0.675&0.691&0.173&0.200&0.715&0.713\\
																				         &&CR&0.709&\textbf{0.751}&\textbf{0.271}&\textbf{0.322}&\textbf{0.719}&\textbf{0.730}\\
																				  \cline{3-9}
					 
					 &\multirow{3}{*}{3k}&RN&0.719&0.693&0.198&0.226&0.689&0.685\\
					 					&&AL&0.691&0.667&0.200&0.229&0.713&0.708\\
					 					&&CR&\textbf{0.756}&\textbf{0.734}&\textbf{0.322}&\textbf{0.340}&\textbf{0.730}&\textbf{0.723}\\

			\midrule
			\multirow{6}{*}{ESP\cite{rhinehart2019precog}} 		  &\multirow{3}{*}{1k}		    &RN&\textbf{0.754}&	\textbf{0.737}&0.123&    0.114&0.700&    0.718\\
																							   &&AL&0.729&	0.717&0.149&	0.156&0.717&	\textbf{0.751}\\
																						       &&CR&0.738&	0.723&\textbf{0.239}&	\textbf{0.243}&\textbf{0.729}&	\textbf{0.751}\\
																				\cline{3-9}
					
																	  &\multirow{3}{*}{3k}       &RN&\textbf{0.797}	&\textbf{0.782} &0.265	&0.290 &0.733	&0.756\\
																								&&AL&0.781	&0.766 &0.216	&0.227 &\textbf{0.764}	&\textbf{0.780}\\
																								&&CR&0.779	&0.764 &\textbf{0.318}	&\textbf{0.340} &0.760	&0.774\\										 
															
													

			\bottomrule
			\end{tabularx}
		\end{threeparttable}
    \caption{Detection results}
    \label{tbl:detection}
\end{table*}

\paragraph{Motion Prediction:}
Table \ref{tbl:prediction} shows the motion forecasting results for various models and curation methods. 
Similar to detection task, we can see that our approach is performing significantly better across the majority of models for all classes. 
Specifically in ILVM model, active learning baseline outperforms our approach in some of the metrics. 
This can be expected as our active learning baseline uses the ILVM backbone for its prediction model. 
This example validates our assumption that even though active learning can be used to select informative examples to improve the performance of a model, the selected training set is not necessarily useful for training other models in the same or different tasks.

Another interesting metric for motion prediction is collision between the predicted trajectories of actors. 
Both ILVM and MultiPath prediction models generate trajectories at scene level, modeling the joint probability distribution among actor trajectories. As shown in the last column of Table \ref{tbl:prediction}, using our curated data, the models are able to generate more consistent predictions for actors.

\begin{table*}[t]
	\centering
		\scriptsize
		\begin{threeparttable}
			\begin{tabularx}{\textwidth}{
				>{\centering\arraybackslash}c |  %
				>{\centering\arraybackslash}X |
				>{\centering\arraybackslash}X |
				>{\centering\arraybackslash}X >{\centering\arraybackslash}X >{\centering\arraybackslash}X>{\centering\arraybackslash}X>{\centering\arraybackslash}X>{\centering\arraybackslash}X>{\centering\arraybackslash}X
				>{\centering\arraybackslash}X >{\centering\arraybackslash}X >{\centering\arraybackslash}X>{\centering\arraybackslash}X>{\centering\arraybackslash}X>{\centering\arraybackslash}X>{\centering\arraybackslash}X}
			\toprule
			Model 	              &  Size   &Train set& \multicolumn{6}{c}{meanADE}  & \multicolumn{6}{c}{minADE}& \multicolumn{2}{c}{Collision @5s}\\
								&					&		& \multicolumn{2}{c}{Vehicles}	  & \multicolumn{2}{c}{Bicycles}& \multicolumn{2}{c}{Pedestrians}& \multicolumn{2}{c}{Vehicles}	  & \multicolumn{2}{c}{Bicycles}& \multicolumn{2}{c}{Pedestrians}  & \multicolumn{2}{c}{Vehicles}\\
								&&&Easy&Hard&Easy&Hard&Easy&Hard&Easy&Hard&Easy&Hard&Easy&Hard&Easy&Hard\\
			\midrule
			\multirow{6}{*}{ILVM\cite{casas2020implicit}} 		  &\multirow{3}{*}{1k}&RN&\textbf{0.67}&0.86&1.07&1.01&1.12&1.18&\textbf{0.49}&0.61&0.87&0.82&0.91&0.95&1.02&1.2\\
																&&AL&0.71&0.90&1.14&0.96&0.90&0.92&0.51&0.65&0.87&0.76&0.77&0.78&0.87&1.12\\  
																&&CR&\textbf{0.67}&\textbf{0.84}&\textbf{0.95}&\textbf{0.83}&\textbf{0.87}&\textbf{0.87}&\textbf{0.49}&\textbf{0.59}&\textbf{0.73}&\textbf{0.66}&\textbf{0.72}&\textbf{0.72}&\textbf{0.67}&\textbf{0.86}\\
																\cline{3-17}
													  
									&\multirow{3}{*}{3k}&RN&\textbf{0.66}&0.86&1.10&0.95&0.93&0.95&0.40&0.50&0.72&0.63&0.66&0.67&0.94&1.10\\
													  &&AL&\textbf{0.66}&\textbf{0.82}&1.05&\textbf{0.83}&\textbf{0.73}&\textbf{0.74}&\textbf{0.34}&\textbf{0.40}&0.62&\textbf{0.52}&\textbf{0.53}&\textbf{0.52}&0.88&1.00\\ 
													  &&CR&0.67&0.87&\textbf{1.0}&0.84&0.74&0.75&0.38&0.48&\textbf{0.61}&0.54&0.54&0.53&\textbf{0.60}&\textbf{0.74}\\
			\midrule
			\multirow{6}{*}{MultiPath\cite{chai2019multipath}} 		  &\multirow{3}{*}{1k}&RN&0.75&0.84&1.22&1.24&1.08&1.08&0.38&0.42&0.74&0.58&0.48&0.41&1.96&2.23\\
																						 &&AL&0.81&0.74&1.47&1.11&1.06&1.00&0.38&0.33&0.68&0.58&0.42&0.37&2.35&1.92\\
																						 &&CR&\textbf{0.71}&\textbf{0.69}&\textbf{0.97}&\textbf{0.93}&\textbf{0.92}&\textbf{0.88}&\textbf{0.36}&\textbf{0.30}&\textbf{0.67}&\textbf{0.45}&\textbf{0.40}&\textbf{0.35}&\textbf{1.73}&\textbf{1.46}\\
																						\cline{3-17}
								 &\multirow{3}{*}{3k}&RN&0.84&1.11&1.24&1.03&1.08&1.09&0.42&0.52&0.58&0.51&0.41&0.41&2.2&2.9\\
													&&AL&0.74&0.94&1.11&0.90&1.00&0.99&0.33&0.41&0.58&0.51&0.37&0.37&1.9&2.2\\
													&&CR&\textbf{0.69}&\textbf{0.89}&\textbf{0.93}&\textbf{0.79}&\textbf{0.88}&\textbf{0.88}&\textbf{0.30}&\textbf{0.37}&\textbf{0.45}&\textbf{0.41}&\textbf{0.35}&\textbf{0.35}&\textbf{1.4}&\textbf{1.8}\\

			\midrule 
			\multirow{6}{*}{ESP\cite{rhinehart2019precog}} 		  
			&\multirow{3}{*}{1k} &RN&1.13	&1.48	&0.40	&1.49	&1.56	&1.56	&0.61	&0.79	&1.01	&0.90	&0.76	&0.79	&2.89	&\textbf{3.80}\\
								&&AL&0.94   &\textbf{1.18}	&0.37	&1.21	&\textbf{1.00}	&\textbf{1.01}	&\textbf{0.43}	&\textbf{0.52}	&0.85	&0.72	&\textbf{0.53}	&\textbf{0.55}	&3.86	&5.00\\
								&&CR&\textbf{0.92}	&1.19	&\textbf{0.28}	&\textbf{1.13}	&1.06	&1.07	&0.48	&0.60	&\textbf{0.73}	&\textbf{0.64}	&0.56	&0.57	&\textbf{3.47}	&4.60\\
							\cline{3-17}
									&\multirow{3}{*}{3k}&RN&1.02	&1.33	&1.50	&1.23	&1.14	&1.17	&0.48	&0.63	&0.81	&0.68	&0.62	&0.63	&3.50	&\textbf{3.74}\\
													   &&AL&\textbf{0.93}	&\textbf{1.21}	&1.59	&1.31	&\textbf{0.96}	&\textbf{0.98}	&\textbf{0.47}	&\textbf{0.60}	&0.96	&0.78	&0.54	&0.55	&\textbf{2.70}	&\textbf{3.74}\\
													   &&CR&0.97	&1.25	&\textbf{1.23}	&\textbf{1.08}	&0.97	&\textbf{0.98}	&0.48	&\textbf{0.60}	&\textbf{0.70}	&\textbf{0.60}	&\textbf{0.52}	&\textbf{0.53}	&3.50	&4.70\\													
            
			\bottomrule
			\end{tabularx}
		\end{threeparttable}
    \caption{Prediction results}
    \label{tbl:prediction}
\end{table*}

\section{Conclusion}
In this paper we presented a dataset curation pipeline for self-driving to select unlabeled data for labeling. We described a set of intuitive complexity measures to characterize the traffic scene of the collected data wrt various aspects including the topology of the map, diversity and complexity of surrounding actors and their behaviors, and the executed maneuver of the SDV. We also presented a method to select interesting and challenging sections of the collected logs using the described measures together with adding diverse examples to the final selected set of scenarios. Through extensive experiments using various models for the main tasks of perception and prediction, we demonstrated the effectiveness of our curation strategy compared to active learning methods. We believe future extensionsof our method will be impactful in areas of robotics where complex perception, prediction and ego-action plays a significant role. 

{\small
\bibliographystyle{ieee_fullname}
\bibliography{references}
}

\appendix
{\noindent \Large \textbf{\Supp} \vspace{0.0cm}} \\

\section{Comparison of Selected Datasets}
\begin{figure}[h]
	\centering
	\includegraphics[width=0.49\textwidth]{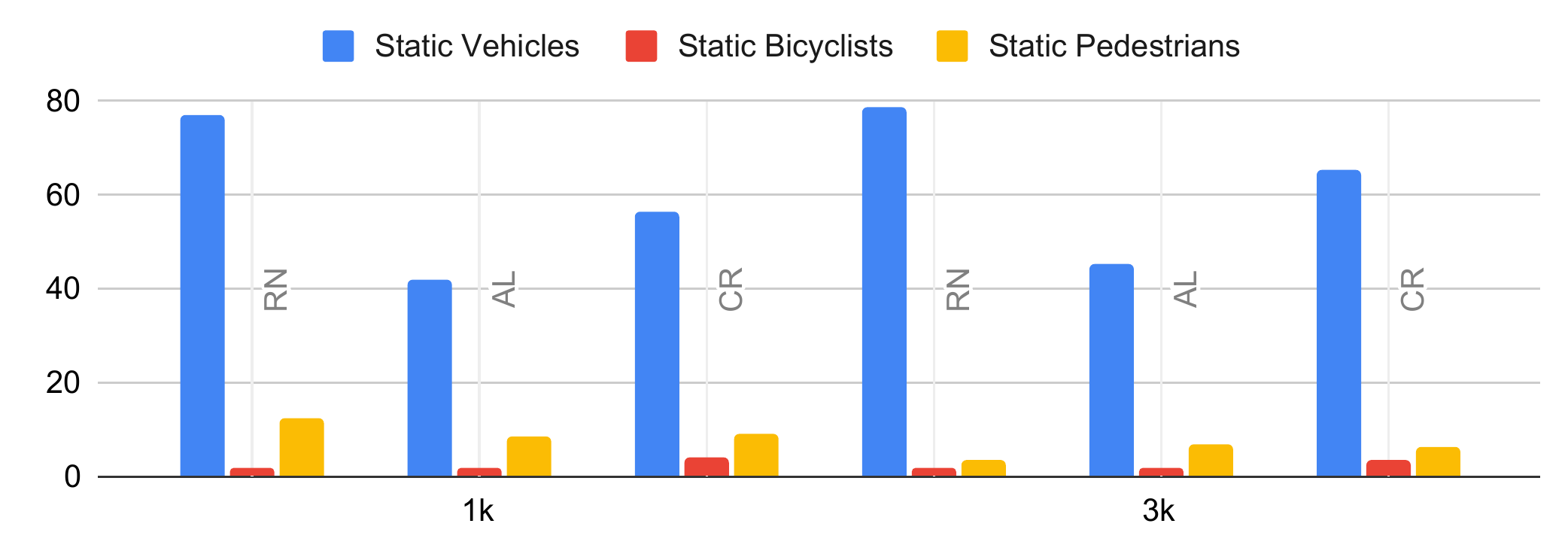}
	\includegraphics[width=0.49\textwidth]{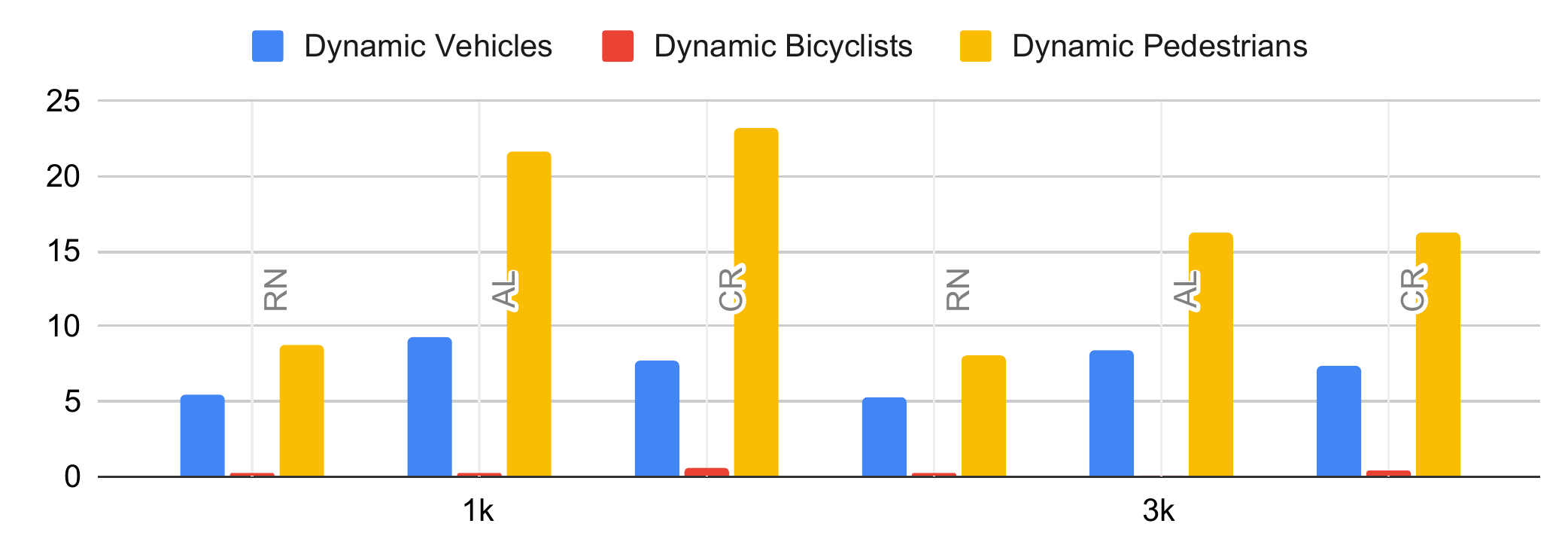}
	\caption[]
    {Label statistics. The plots show mean number of actor class over all the frames in the dataset selected by random sampling (RN), active learning baseline (AL), and out proposed curation method (CR).}
    \label{fig:stats}
\end{figure}

Figure \ref{fig:stats} shows the mean number of static\footnote{We consider actors with less than 0.5$\frac{m}{s}$ speed as static.} and dynamic actors available in the selected datasets by the baseline methods and oru proposed approach. 
This indicates that the random sampling (RN) baseline, as expected, selects scenes that has much more static actors, as the majority of the actors in the snippets are likely to be stopped. 
On the other hand, both active learning (AL) and our curation method (CR) are able to pick snippets with significantly more dynamic actors. 
Note that AL selects selects scenes with marginally more dynamic vehicles in both 1k and 3k settings compared to CR. 
For dynamic pedestrians, the sets  selected by AL and CR have equal number of labels in 3k, with CR having marginally more pedestrians in 1k setting. Finally, CR selects snippets with significantly more number of Bicycles. 
Additionally, even though in AL and CR have close label statistics for dynamic vehicles and pedestrians, our approach yields to significantly better perception and prediction results as discussed in the main paper.

\section{Qualitative Examples of Complexity Measures}
\begin{figure*}[h]
	\centering
	\includegraphics[width=0.77\textwidth]{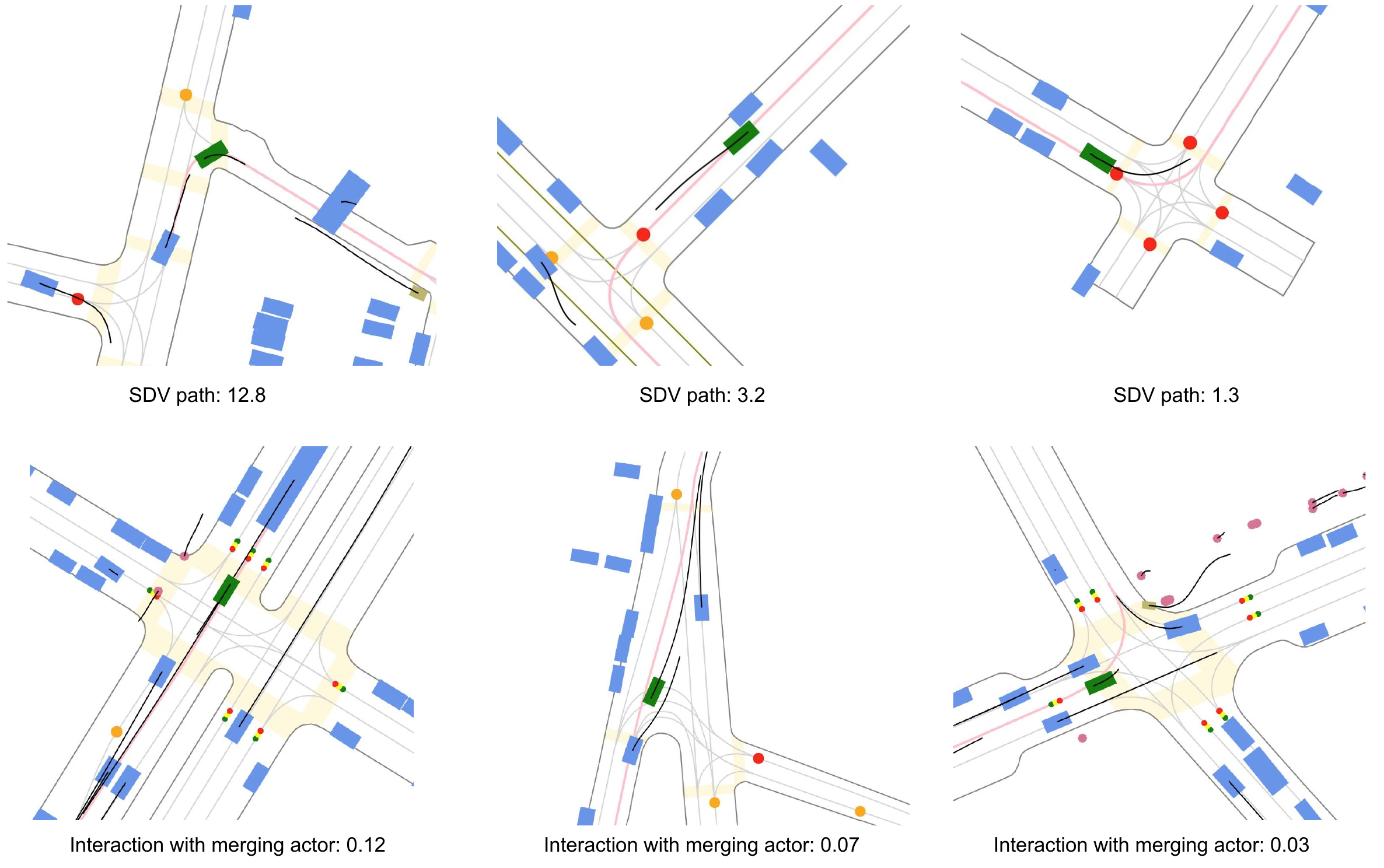}
	\caption[]
    {Qualitative examples of complexity measures related to SDV path (top row) and the interactions between SDV and actors that are merging into SDV's lane (bottom row).}
    \label{fig:sdv}
\end{figure*}

\begin{figure*}[h]
	\centering
	\includegraphics[width=0.77\textwidth]{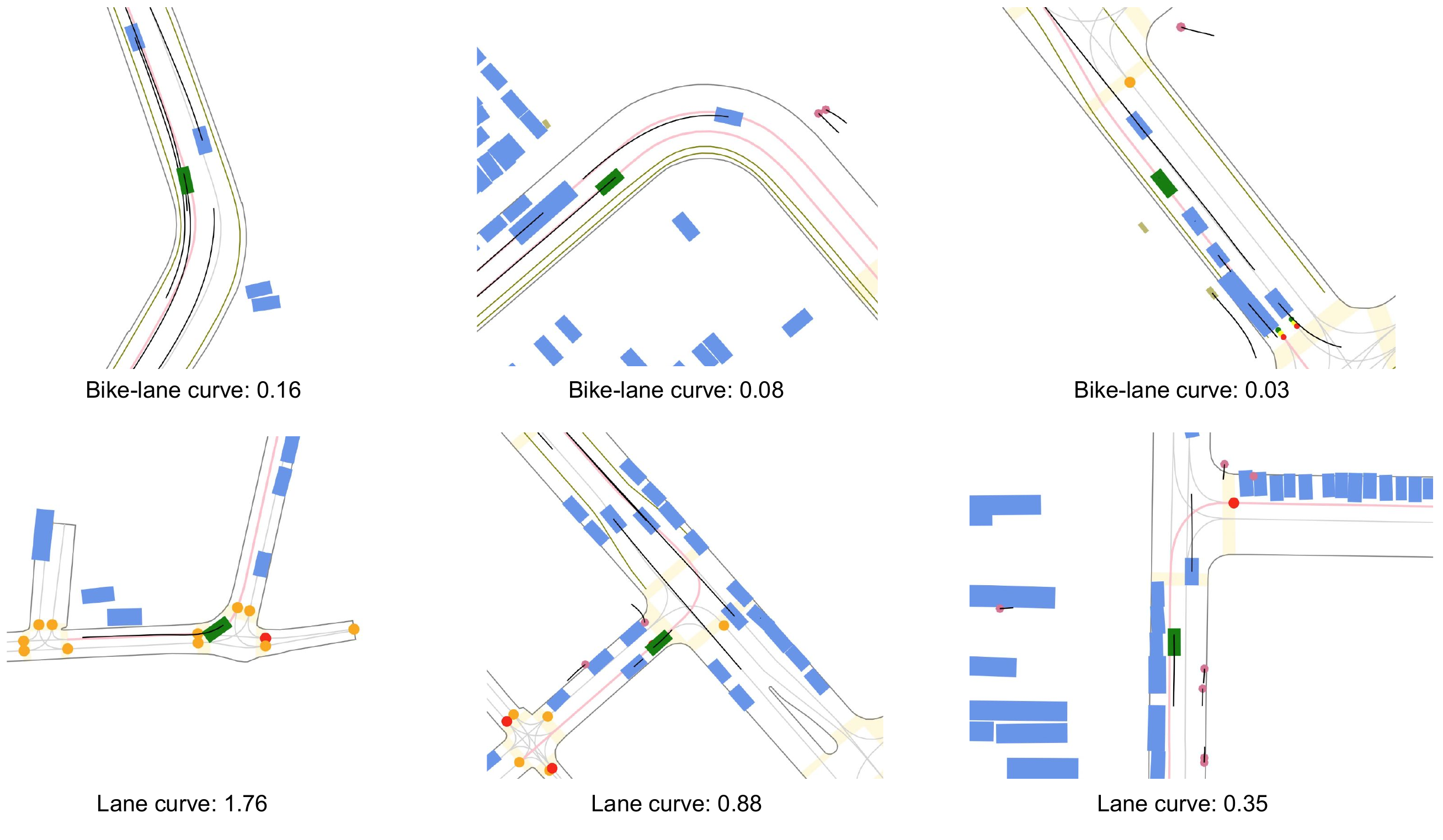}
	\caption[]
    {Qualitative examples of complexity measures related to vehicle (top row) and bike (bottom row) motion-paths.}
    \label{fig:lane}
\end{figure*}

\begin{figure*}[h]
	\centering
	\includegraphics[width=1.0\textwidth]{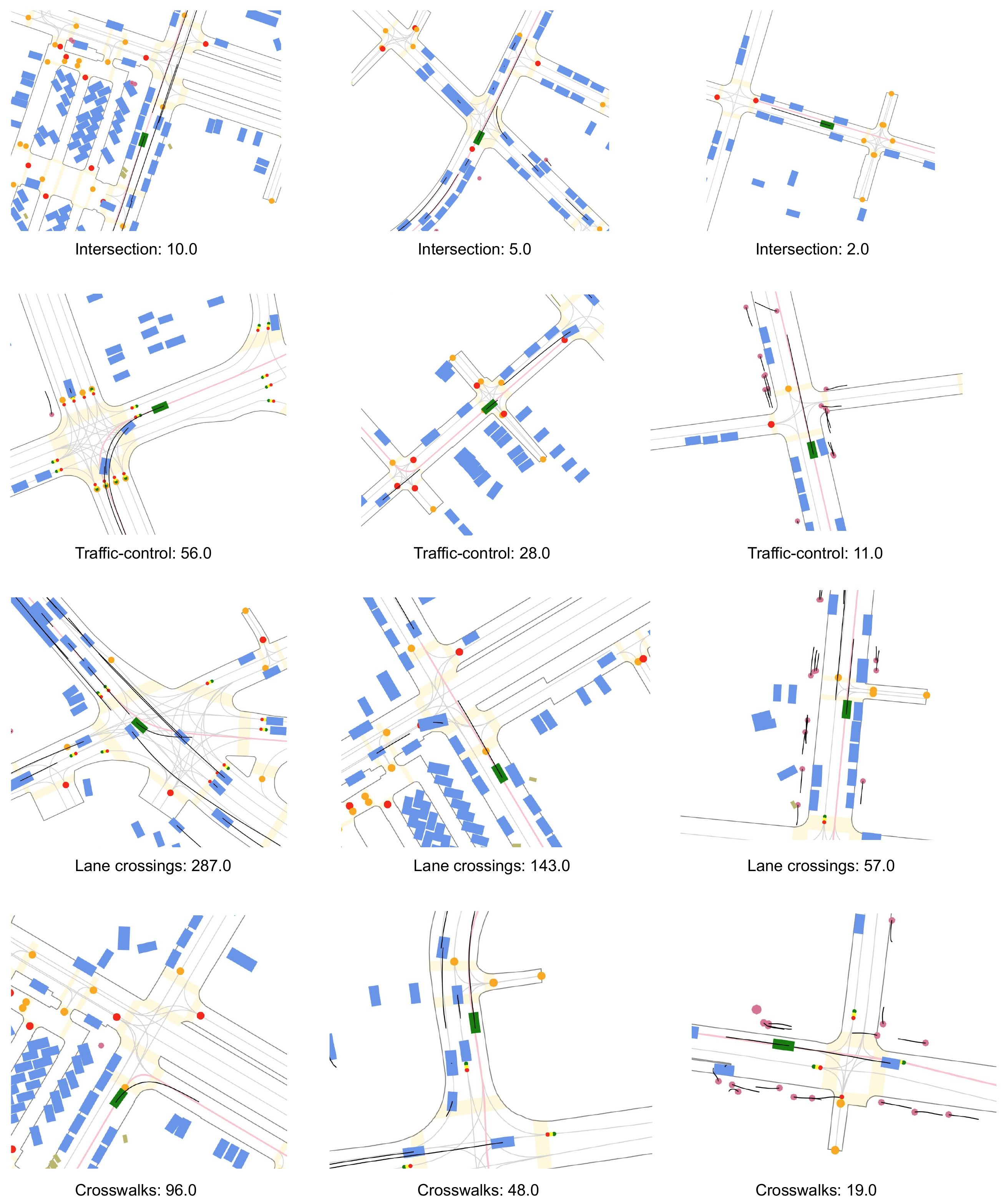}
	\caption[]
    {Qualitative examples of complexity measures related to map.}
    \label{fig:map}
\end{figure*}

\begin{figure*}[h]
	\centering
	\includegraphics[width=1.0\textwidth]{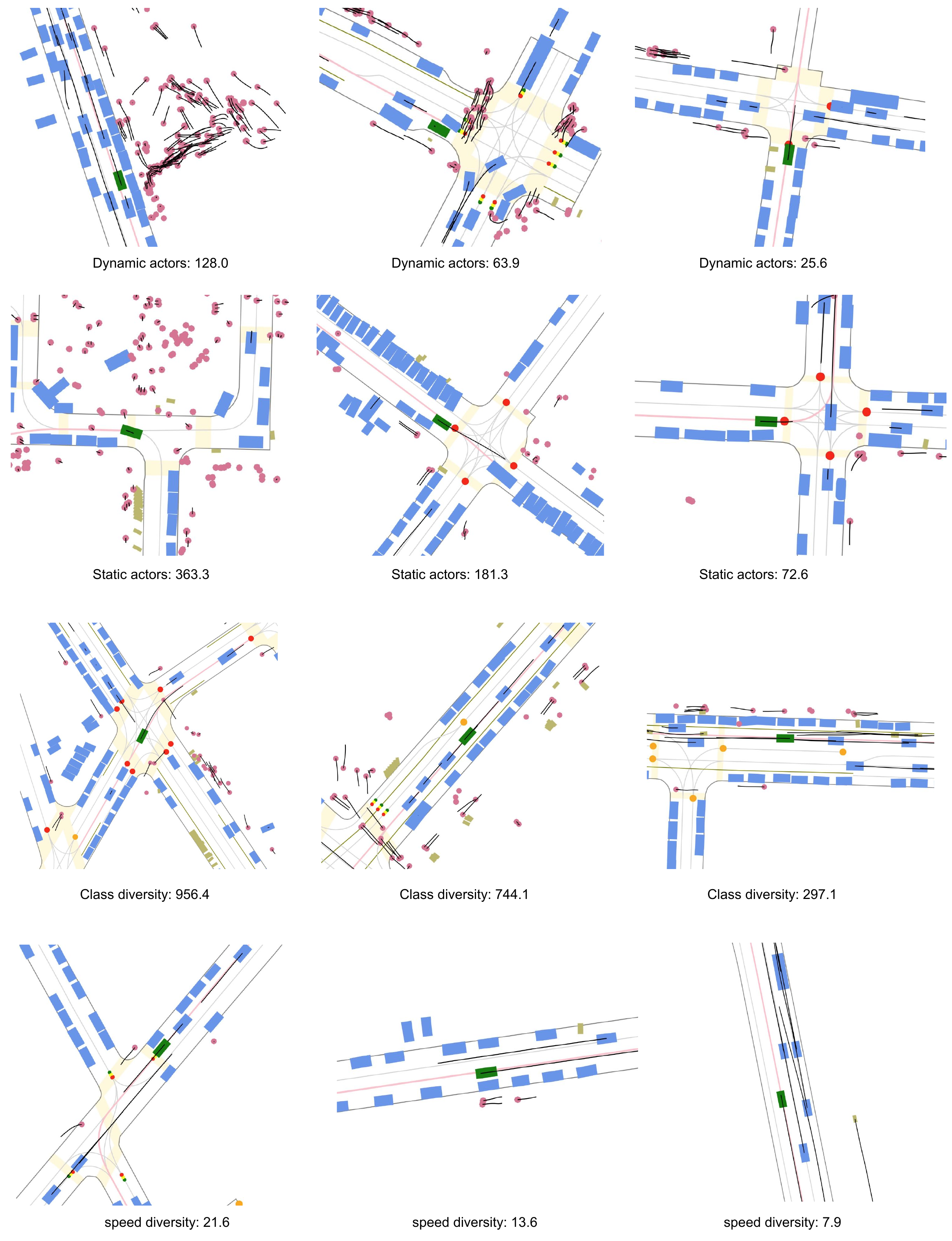}
	\caption[]
    {Qualitative examples of complexity measures related to actors in the traffic-scene.}
    \label{fig:actors}
\end{figure*}

Figures \ref{fig:sdv}-\ref{fig:actors} present qualitative examples of the our proposed complexity measures. Each row provides examples for a specific complexity measure, decreasing from left to right. 
Specifically, in Figure \ref{fig:sdv}, examples of SDV path is shown. In the top-left scene, the SDV is performing a very sharp turn and hence high SDV-path complexity measure. 
Similarly, the the top-middle figure, the SDV performs a nudging maneuver which results in a path with some curvature and varying curvature rate of change, compared to the top-right scene which has a fixed curvature. 
On the bottom row, examples of interaction complexity measures are presented. 
In bottom-left scene, another vehicle is merging into SDV lane, with SDV and the other vehicle having high speed. The bottom-middle and right scenes, show similar interactions where the time-to-interaction is higher and less interesting.
In Figure \ref{fig:map} bottom row, the crosswalk measure is shown where the number of motion paths that intersect with the crosswalk polygon indicate the complexity. 
Similarly, The lane-crossings, shown on the second row from bottom, is directly correlated to how complex the traffic situation can be. 
In Figure \ref{fig:lane}, the lane-curve complexity is shown. 
Note that the curvature of the lanes in bottom-left scene is high as the turns are sharp, compared to the other scenes there they are more open.

\end{document}